\documentclass[runningheads]{llncs}

\usepackage{eccv}

\usepackage{eccvabbrv}

\usepackage{placeins}
\usepackage{graphicx}
\usepackage{booktabs}
\usepackage{multirow}

\usepackage[accsupp]{axessibility}  

\usepackage{hyperref}

\usepackage{orcidlink}

\begin{document}

\title{$\mu$Match: Foundation Models for Semi-supervised Learning and Domain Adaptation in EM} 

\titlerunning{$\mu$Match: Foundation Models for SSL and DA in EM}

\author{Marei Freitag\inst{1,2}\orcidlink{0009-0005-0920-2535} \and 
Olesia Korchevaia\inst{2}\orcidlink{0009-0007-8475-9637} \and
Luca Freckmann\inst{2}\orcidlink{0000-0002-8285-2586} \and
Anwai Archit\inst{2}\orcidlink{0009-0002-9533-8620} \and 
Constantin Pape\inst{2}\orcidlink{0000-0001-6562-7187}}

\authorrunning{M.~Freitag et al.}

\institute{Life and Medical Sciences Institute (LIMES), University of Bonn, Germany \and 
Institute of Computer Science, Georg-August-University Göttingen, Germany}

\maketitle

\begin{abstract}
  Vision foundation models have substantially advanced computer vision, enabling state-of-the-art performance in zero- and few-shot settings. They have been successfully applied to biomedical imaging tasks ranging from organ segmentation in computed tomography to cell segmentation in light microscopy. Electron microscopy (EM) is a central modality for analyzing cellular ultrastructure due to its nanometer-scale resolution. However, the application of foundation models in EM has so far been limited to specific organelles, such as mitochondria, largely due to the diversity of segmentation tasks and the scarcity of comprehensively annotated data. As a result, EM segmentation still predominantly relies on supervised learning, requiring extensive manual annotation and limiting ultrastructural analysis. To address this gap, we propose $\mu$Match, a framework for semi-supervised learning and domain adaptation that leverages foundation models. We implement state-of-the-art student–teacher-based methods and evaluate multiple foundation models (SAM, SAM2, $\mu$SAM, DINOv2/v3) on challenging EM tasks, including mitochondrion, nucleus, and neurite segmentation. Our results demonstrate consistent improvements over strong baselines and highlight a path toward substantially reducing the annotation effort in EM.
\end{abstract}

\section{Introduction} \label{sec:intro}
Biomedical image segmentation has undergone substantial advances, initially driven by supervised deep learning and more recently by vision foundation models (VFMs). These models enable zero- or few-shot segmentation in medical imaging \cite{ma2024segment, archit2025medicosam, totalsegmentator}, fluorescence microscopy \cite{archit_segment_2025, cellpose-sam}, and histopathology \cite{patho-sam, cellvit}. Another key imaging modality is electron microscopy (EM), which enables the study of ultrastructure at cellular \cite{openorganelle}, tissue and organ \cite{liver-em, microns, fafb-seg}, and organism \cite{platy} scales. However, EM has not yet benefited from foundation models to the same extent as other modalities, largely due to the high diversity of segmentation tasks, particularly across different organelles, and the limited availability of annotated data. While initial efforts for mitochondria and nuclei exist \cite{mitonet, archit_segment_2025}, a comprehensive foundation model for organelles remains unavailable and would require substantial additional data collection and annotation.

\begin{figure}[h!]
  \centering
  \includegraphics[width=0.99\linewidth]{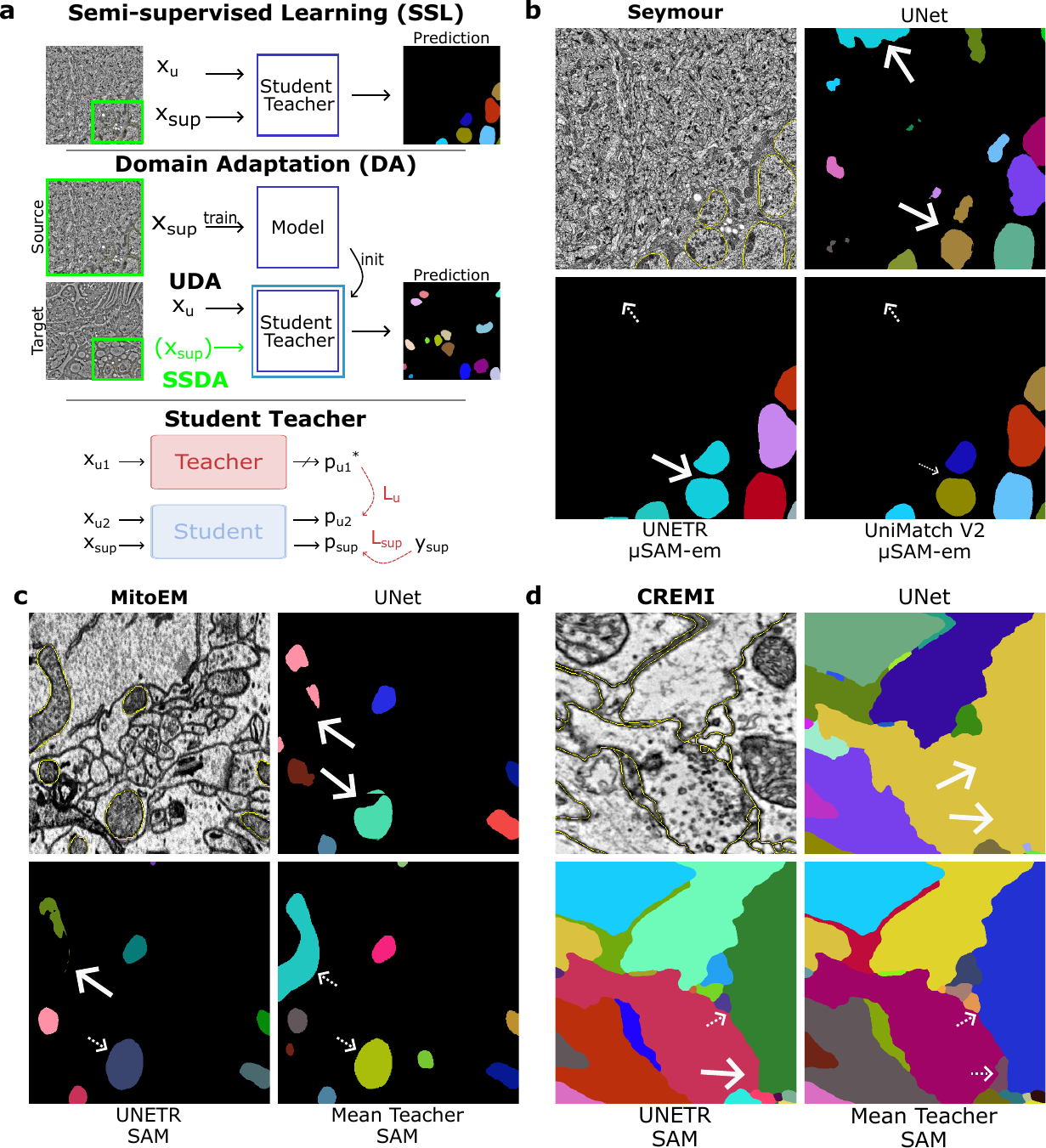}
  \caption{$\mu$Match overview: \textbf{a)} we study \textbf{SSL}, training on data with a small labeled part (green box), and \textbf{DA}, pre-training on a labeled source followed by training on the unlabeled (UDA) or partially labeled (SSDA) target. The \textbf{student teacher} uses augmented views ($x_{u1}$, $x_{u2}$) to compute the unsupervised loss $L_u$, and a supervised loss $L_{sup}$ on labeled samples. We implement three different methods: Mean Teacher, FixMatch, and UniMatch v2, and initialize the models with foundation model weights. \textbf{b) - d)} Exemplary results from the UNet baseline, the best supervised model, and a corresponding SSL model, for nuclei (Seymour), mitochondria (MitoEM), and neurites (CREMI). Bold arrows: segmentation errors; dashed arrows: correct segmentation.}
  \label{fig:overview}
\end{figure}

In the absence of a comprehensive foundation model for EM, segmentation tasks are predominantly addressed using supervised learning, which entails substantial manual annotation effort and limits the scalable quantification of cellular ultrastructure. To mitigate this limitation, semi-supervised learning (SSL) and domain adaptation (DA) have been explored in EM, employing both domain alignment strategies \cite{roels_domain_2018, bermudez-chacon_visual_2020, matskevych_shallow_2022, januszewski2019segmentation} and student–teacher-based approaches \cite{huang_semi-supervised_2022, rumberger_actis_2023, muth_synapsenet_2025, archit_probabilistic_2023}.
Recent advances in computer vision have unified student–teacher-based SSL and DA \cite{berthelot_adamatch_2022} and incorporated foundation models to achieve state-of-the-art performance in semi-supervised segmentation \cite{yang_unimatch_2025}. However, these developments have not yet been systematically translated to EM, despite the clear need for methods that reduce the annotation burden.

Here, we propose $\mu$Match, a unified framework for SSL and DA in EM that leverages VFMs. See Fig.~\ref{fig:overview} for a contribution overview. Specifically, we:
\begin{itemize}
    \item Systematically evaluate multiple foundation models (SAM, SAM2, $\mu$SAM, DINOv2/v3) in supervised, SSL, and DA settings. 
    \item Implement and assess three student–teacher-based methods for SSL and DA in EM: Mean Teacher \cite{tarvainen_mean_2018}, FixMatch \cite{sohn_fixmatch_2020}, and UniMatch v2 \cite{yang_unimatch_2025}.
    \item Evaluate these approaches on a diverse set of datasets and segmentation tasks, including nucleus, mitochondrion, and neurite segmentation.
    \item Conduct an ablation study on the confidence threshold used for pseudo-labeling, a key hyperparameter in $\mu$Match.
\end{itemize}

\section{Related Work} 

Segmentation tasks in EM can broadly be divided into two categories: segmentation of cellular compartments, such as complete cells or neural processes, and segmentation of organelles or other ultrastructural components. Compartment segmentation typically corresponds to an instance segmentation problem, whereas organelle segmentation can be formulated as either instance or semantic segmentation, depending on the organelle of interest. Here, we address nucleus, mitochondrion, and neurite segmentation. Nucleus \cite{spiers2021deep, machireddy2023segmentation, platy} and mitochondrion \cite{xiao2018automatic, franco2022stable, li2022advanced} segmentation are commonly addressed using a UNet that predicts the foreground and often additional outputs to separate instances. In addition, two models, MitoNet \cite{mitonet} and $\mu$SAM \cite{archit_segment_2025}, have been trained on large and diverse datasets for mitochondrion segmentation and, in the case of $\mu$SAM, also nucleus segmentation, and have demonstrated robustness across varying conditions. Neurite segmentation poses additional challenges due to the complex morphology, topology, and spatial extent of neural processes. It is typically approached via boundary or affinity prediction followed by clustering \cite{beier2017multicut, lee2021learning, funke2018large}, or with flood-filling networks \cite{januszewski2018high}, which segment individual neurites iteratively. Multi-organelle segmentation, including structures such as endoplasmic reticulum, Golgi apparatus, and endosomes, in addition to mitochondria and nuclei, is commonly formulated as a semantic segmentation task \cite{openorganelle, gallusser2022deep}. In practice, these tasks are predominantly addressed using supervised learning, as reflected in the references cited above. Approaches for SSL and DA in EM are discussed below.

Data annotation for segmentation in EM is extremely labor-intensive, as it requires dense pixel-wise labels generated by domain experts. Consequently, leveraging unlabeled data through SSL or transferring knowledge from labeled source datasets via DA is highly desirable. Various strategies for SSL and DA have been proposed. Here, we focus on student–teacher approaches, which are applicable to both settings. In these methods, a student model is trained to match the predictions of a teacher, typically using different augmented views of the same input, while the teacher’s weights are derived from the student. In SSL, an additional supervised loss is applied to labeled samples, whereas in unsupervised DA (UDA) the model is initialized with weights from training on the source domain. Semi-supervised DA (SSDA) combines initialization with a source model and an SSL objective in the target domain. See Fig.~\ref{fig:models} for an overview of these settings. Alternatively, DA can be performed by jointly training on labeled source data and unlabeled or partially labeled target data. However, we do not consider this setting due to the practical limitation of requiring access to all source data during training.

The Mean Teacher \cite{tarvainen_mean_2018} was among the earliest student–teacher approaches for SSL. It applies weak augmentations to generate two different views for the student and teacher models, respectively, and updates the teacher using an exponential moving average (EMA) of the student’s weights. FixMatch \cite{sohn_fixmatch_2020} simplifies this setup by sharing the weights of student and teacher, preventing gradient updates in the teacher. It further introduces a combination of weak and strong augmentations and a confidence threshold for pseudo-labeling, discarding predictions below this threshold from the loss. AdaMatch \cite{berthelot_adamatch_2022} extends this paradigm to DA by jointly processing samples from both source and target domains. It further dynamically adjusts the confidence threshold based on performance on weakly augmented source data. The methods discussed so far were applied to classification. UniMatch \cite{yang_revisiting_2023} adapts FixMatch to semantic segmentation and enhances it through feature perturbations and the use of two strongly augmented views. UniMatch v2 \cite{yang_unimatch_2025} streamlines this architecture and replaces the commonly used ResNet encoder with the pre-trained Vision Transformer of DINOv2, resulting in significantly improved segmentation performance.

Prior work has investigated SSL and DA for segmentation in EM. Several approaches to SSL rely on feature alignment strategies rather than student–teacher frameworks. Y-Net \cite{roels_domain_2018}, for example, introduces an additional decoder that performs image reconstruction in the target domain to encourage alignment between source and target features, and evaluates this strategy for mitochondrion segmentation. The authors of \cite{januszewski2019segmentation} combine a flood-filling network \cite{januszewski2018high} with a CycleGAN \cite{zhu2017unpaired} to adapt the visual style of target images to the source domain. Other approaches leverage self-supervised learning to promote feature alignment \cite{bermudez-chacon_visual_2020} or incorporate intermediate predictions from feature-based machine learning \cite{matskevych_shallow_2022}.

More recent work has adopted student–teacher approaches for both SSL and DA in EM segmentation. WDA-Net \cite{qiu_wda-net_2022} uses multi-task learning for joint point detection, object detection, and instance segmentation of mitochondria, using task-specific decoders to provide mutual supervision signals. It was later extended to incorporate a pseudo-label filtering mechanism \cite{qiu_weakly-supervised_2024}. Other studies adapt the Mean Teacher to EM, for example by introducing reconstruction-based pretraining to improve neuron segmentation \cite{huang_semi-supervised_2022}. SynapseNet \cite{muth_synapsenet_2025} integrates Mean Teacher-based DA into a framework for segmenting synaptic structures in EM images. Another line of work focuses on improving confidence estimation for pseudo-labeling: ACTIS \cite{rumberger_actis_2023} derives uncertainty estimates from multiple augmented views, although it is evaluated on light microscopy rather than EM. Similarly, \cite{archit_probabilistic_2023} use a probabilistic UNet \cite{kohl2018probabilistic} to obtain calibrated uncertainties and evaluates the approach on mitochondrion segmentation, among other tasks.

Unlike prior work in EM, we leverage VFMs to initialize the encoder of the segmentation network, motivated by the substantial performance gains demonstrated for natural images in \cite{yang_unimatch_2025}. We implement EM-specific variants of Mean Teacher, FixMatch, and UniMatch v2 (see Fig.~\ref{fig:models}) and systematically evaluate them across a diverse set of SSL and DA scenarios and segmentation tasks.

\section{Methods}
We provide an overview of our methodology for supervised learning (Sec.~\ref{sec:supervised}), SSL (Sec.~\ref{sec:semisup}), and DA (Sec.~\ref{sec:da}); as well as the evaluation metrics (Sec.~\ref{sec:metrics}).

\subsection{Supervised training} \label{sec:supervised}

Supervised training is performed using labeled data block(s) for training and for validation. We train UNet models from scratch and train a UNETR model \cite{hatamizadeh2022unetr} whose encoders are initialized with the weights of either SAM, SAM2, DINOv2, or DINOv3. In addition, we also initialize with the domain-specific VFMs $\mu$SAM-lm and $\mu$SAM-em, which adapt SAM to cell segmentation in light microscopy and mitochondrion / nucleus segmentation in EM, respectively \cite{archit_segment_2025}. The input is normalised to a range of $[0, 1]$ for UNet, SAM2 and DINOv2/v3 and to a range from $[0, 255]$ for all other SAM based models. Each model is trained for 10,000 iterations. For the UNETR a smaller patch shape than for UNet is used due to memory constraints, reduced from 512 $\times$ 512 $\times$ 24 voxels to 512 $\times$ 512 $\times$ 4. For each setting, the best base model is selected independently to ensure that each architecture is evaluated under its own optimal conditions. The specific training parameters for each setting are summarized in Tab.~\ref{tab:training_parameters} (Appendix). 
For all experiments with boundary outputs, flip augmentations are applied in each dimension. For affinity outputs, only intensity augmentations are used to avoid corrupting the affinities. Specifically, we use random Gaussian blur with $\sigma$ ranging from 0.1 to 1.0 and random additive Gaussian noise with mean 0 and standard deviation 0.3.  The Adam optimizer is used and a learning rate scheduler with an initial learning rate of 1e-4 that decreases by a factor of 0.5 when the validation loss plateaus is used.

\begin{figure}[h]
  \centering
  \includegraphics[width=\linewidth]{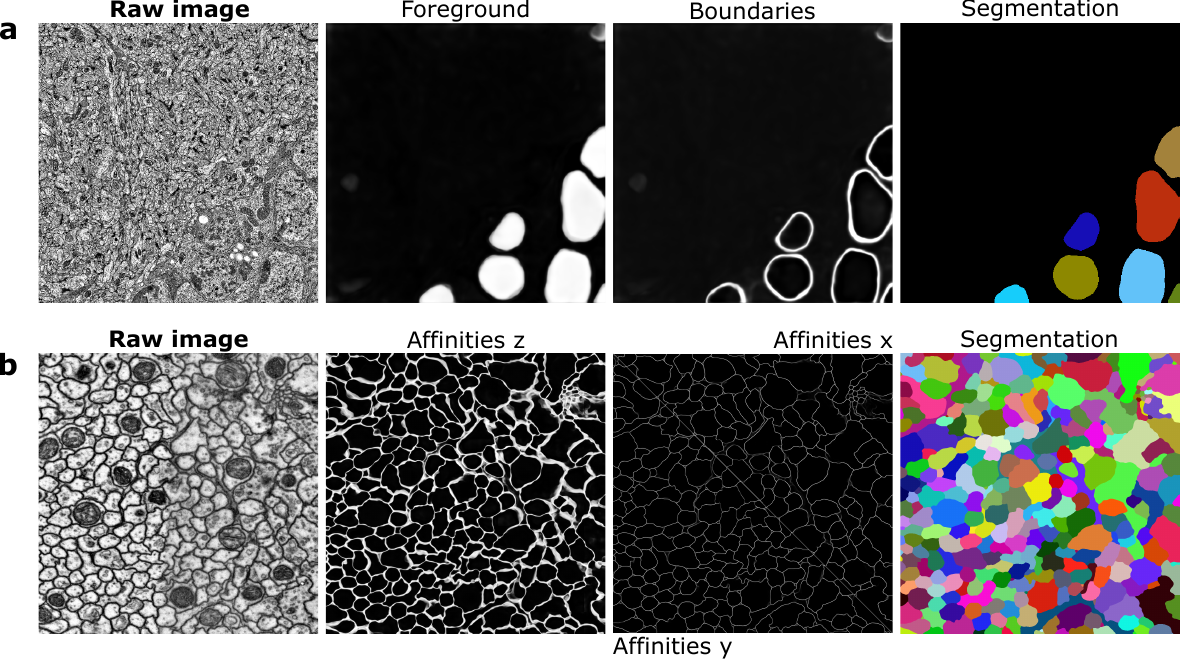}
  \caption{\textbf{a)} Exemplary foreground, boundary predictions, and derived segmentation for nucleus (shown here) and mitochondrion segmentation. \textbf{b)} affinity predictions and derived segmentation for neurites.}
  \label{fig:model_outputs}
\end{figure}

The number of output channels depends on the respective segmentation task. For nucleus and mitochondrion segmentation, we predict two channels corresponding to foreground and boundary probabilities. Instances are computed from these predictions by deriving connected components of the foreground, which serve as seeds for a watershed on the boundaries. The segmentation is then size filtered to retain objects larger than 10,000 voxels.
For neurite segmentation, the model predicts three output channels representing affinities along the x, y and z directions. The instance segmentation is derived by first computing an over-segmentation via watershed, from which a region graph is derived. Edge strengths are then computed based on affinities aggregated over region borders and these strengths are used as costs for a Multicut graph partition, which is translated to the instance segmentation. This approach is adopted from \cite{beier2017multicut}.  
The bias parameter $\beta$ in the cost computation is selected individually for each model via a grid search from 0.5 to 0.9 (step size 0.05) on the validation block(s). Example outputs and derived instances for the two set-ups are shown in Fig.~\ref{fig:model_outputs}.

\subsection{Semi-supervised training} \label{sec:semisup}

We extend training to a semi-supervised set-up, using the same labeled data block(s) as in the supervised setting together with additional unlabeled data. We use the same model outputs, instance segmentation procedures, and hyperparameters as in Sec.~\ref{sec:supervised}, unless stated otherwise.

\begin{figure}[h]
  \centering
  \includegraphics[width=\linewidth]{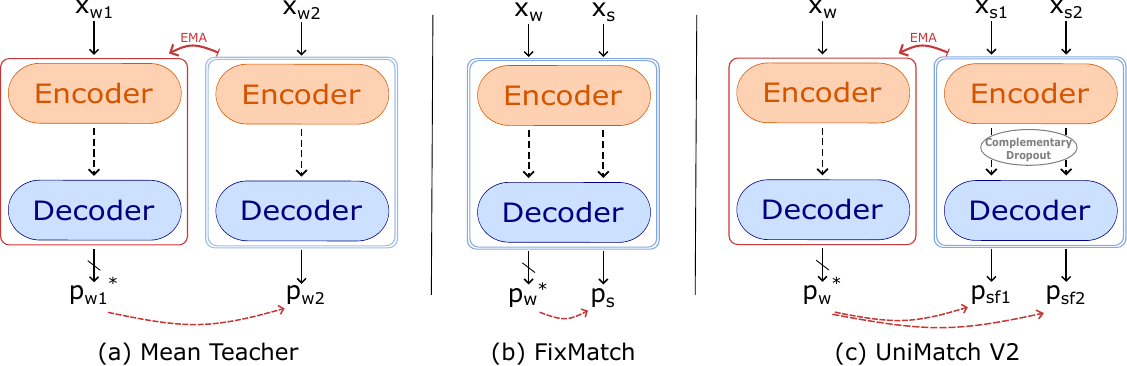}
  \caption{Comparison of the three teacher student methods (\textbf{a) - c)}). $x_w$ represents weakly augmented views, $x_s$ strongly augmented views, $p$ the corresponding prediction. $*$ indicates confidence thresholding, $p_f$ predictions after feature dropout.}
  \label{fig:models}
\end{figure}

We implement three different student-teacher methods for SSL: Mean Teacher, FixMatch, and UniMatch v2. An overview of these methods is shown in Fig.~\ref{fig:models}. In Mean Teacher, weak augmentations are applied to the unlabeled input images. The student receives one weakly augmented view and the teacher the other weakly augmented view. Pseudo-labels are derived from the outputs of the teacher model by applying the \emph{confidence threshold} $t_c$. Only the pixels where the prediction probability is higher than $t_c$ or lower than $1 - t_c$ are used for training, others are masked in loss and gradients. The unsupervised loss is then computed as the Dice loss of student and teacher prediction, subject to the masking procedure we just described. In the supervised pass, the student receives the input. Its prediction and the ground-truth labels are used to determine the supervised loss, for which we use the Dice loss. The total loss 
\begin{align*}
    L = (L_{sup} + L_u) / 2
\end{align*}
is then used to update the student’s weights via backpropagation. Subsequently, the weights of the teacher are updated via exponential moving average (EMA) of the student’s weights.

The FixMatch set-up is similar to Mean Teacher. However, a weakly augmented view is fed to the teacher and a strongly augmented view to the student. The teacher is a continuously updated copy of the student, gradients are not propagated to the teacher. Confidence thresholding, loss computation etc. are the same as in Mean Teacher.

UniMatch v2 uses a weakly augmented view as input to the teacher and two strongly augmented views for the student. The student first computes the supervised prediction, which is backpropagated directly by comparing the prediction with its supervised label to reduce memory consumption. Subsequently, it processes both strongly augmented views. Then, complementary dropout is applied to its features with a ratio of 0.5: for one prediction, 50\% of the feature map channels are set to zero, while for the other predictions, the remaining feature map channels are zeroed out. To preserve the overall magnitude, the feature values are multiplied by a factor of 2. The teacher's weights are then updated via EMA. Besides these differences, confidence thresholding, loss computation etc. are the same as in Mean Teacher and FixMatch.

For the experiments with boundary output channel, the weak data augmentations mentioned consist of random horizontal and vertical flips as well as 90-degree rotations. These geometric augmentations are inverted after the prediction. Strong augmentations additionally include intensity augmentations: random Gaussian blur with kernel size 3 $\times$ 3 and $\sigma$ ranging from 0.1 to 1.0 and random Gaussian noise with mean 0 and standard deviation 0.1. 
For experiments with affinity output, we only use intensity augmentations. For the weak augmentations, these are specifically random Gaussian blur with kernel size 3 $\times$ 3 and $\sigma$ ranging from 0.5 to 1.5, random brightness with a factor ranging from 0.9 to 1.1 and random Gaussian noise with mean 0 and standard deviation 0.2. For the strong augmentations, the parameters are increased: random Gaussian blur with kernel size 5 $\times$ 5 and $\sigma$ ranging from 1.0 to 3.0, random brightness with a factor ranging from 0.6 to 1.4 and random Gaussian noise with mean 0 and standard deviation 0.5.

The pixels within the confidence mask $t_c$ critically determine the success of SSL. On the one hand, if there are too few pixels, the gradients are too sparse and learning is hindered. On the other hand, if there are many "low-quality" pixels, the training is also negatively affected.
To alleviate this issue, we train the model with an initial supervised warm-up phase of 1,000 iterations to ensure that the teacher predictions are already meaningful. 
Furthermore, we explore different settings for the confidence threshold $t_c$. First, setting it to a fixed value throughout the entire training process. Here, we evaluate the thresholds $t_c = 0.5, 0.7, 0.9$.
Note that $t_c = 0.5$ corresponds to not applying a mask.
Alternatively, the confidence threshold can be updated during training. For this purpose, we use a confidence threshold scheduler with two different possible strategies:
decreasing $t_c$ when the training loss stagnates, or increasing $t_c$ monotonically over the course of training. In both settings, a warm-up with fixed $t_c = 0.5$ can be defined. 
Increasing $t_c$ monotonically is used in our experiments unless stated otherwise.

\subsection{Domain Adaptation} \label{sec:da}

For DA, supervised training is performed on the (fully labeled) source domain for 25,000 iterations, which replaces the warm-up supervised pre-training on the target data. Afterwards, a student–teacher is trained either in an unsupervised setting (UDA) where only unlabeled data and the unsupervised loss is used, or in a semi-supervised setting (SSDA) where a small fraction of the target data with labels is also used for training. The difference between SSL, UDA, and SSDA is also illustrated in Fig.~\ref{fig:overview} a).

\subsection{Metrics} \label{sec:metrics}

Results are evaluated using task-specific instance segmentation metrics. For mitochondria and nuclei, the mean segmentation accuracy (mSA) \cite{hirling2024segmentation} is used. It represents the ratio of correctly predicted instances to the total number of predicted and ground truth instances, evaluated across a range of overlap thresholds between 0.5 and 0.95, with a step size of 0.05. Higher mSA values indicate better segmentation performance.
For neurite segmentation, performance is evaluated using the score defined by the CREMI challenge \cite{cremi_website} (CREMI Score), a composite metric that evaluates instance segmentation quality by focusing on topological correctness rather than pixel-wise overlap. A detailed description of the metrics calculation can be found in App.~\ref{sec:appendix_metrics}. 

\section{Results}

\subsection{Data}\label{sec:data}

We evaluate nucleus segmentation on a serial section transmission EM (ssTEM) dataset from the brain of a first instar \textit{Drosophila} larva \cite{winding2023seymour} (Seymour) with voxel size 32 $\times$ 32 $\times$ 50 nm (x $\times$ y $\times$ z). 
We use 6 blocks with nucleus labels for supervised and SSL experiments, 1 for training, 1 for validation, 4 as test, and an additional 15 blocks without labels for SSL. Each block measures 512 $\times$ 512 $\times$ 64 voxels. For data scaling experiments (App.~\ref{sec:appendix_data_scaling}), we use up to 3 additional blocks with labels.
The nuclei in this data are annotated by us in $\mu$SAM via interactive instance segmentation.
For DA, we use the FAFB dataset \cite{mu2021fafb}, which is derived from a ssTEM volume of the female adult \textit{Drosophila} brain with voxel size 32 $\times$ 32 $\times$ 40 nm. Here, we use 6 blocks with labels (1 train, 1 val, 4 test), which measure 512 $\times$ 512 $\times$ 64. In addition, we use 10 blocks of size 512 $\times$ 512 $\times$ 256 without labels (8 train, 2 val).
We also use a dataset derived from MICrONS \cite{microns}, which contains ssTEM data of murine cortex and nucleus annotations, at a voxel size of 64 $\times$ 64 $\times$ 40 nm. We again use 6 labeled blocks (1 train, 1 val, 4 test) and 10 unlabeled blocks (8 train, 2 val). Each measures 512 $\times$ 512 $\times$ 512 voxels.

For mitochondrion segmentation, we use the MitoEM dataset \cite{wei2020mitoem}, which consists of two large EM volumes of approximately 30 $\mu m^3$ at a voxel size of 8 $\times$ 8 $\times$ 30 nm, derived from human and rat cortex. We use 4 labeled blocks in supervised and SSL experiments, 2 for training and 2 for validation (one human, one rat per split). Each block measures 512 $\times$ 512 $\times$ 64 voxels. We use large unlabeled blocks for SSL: of size 4096 $\times$ 4096 $\times$ 336  for training and of size 4096 $\times$ 4096 $\times$ 64 for validation, one for human and rat for each split. Evaluation is performed on two separate test blocks (human and rat) of size 4096 $\times$ 4096 $\times$ 100.
For DA, we use the MitoEM v2 dataset \cite{liu2025mitoemv2}, selecting 6 of its 8 available conditions. The Mossy and Pyra conditions are excluded because they are only sparsely labeled, preventing evaluation with our metrics. For each selected dataset, the data are divided into splits as specified in Tab.~\ref{tab:me2_datasets} (Appendix). One block (512 $\times$ 512 $\times$ 64 voxels) with labels is used for training and validation, respectively.

We use the CREMI \cite{cremi_website} dataset for neurite segmentation. It contains individual annotated neurites in EM data extracted from the fruit fly brain. The data has a voxel size of 4 $\times$ 4 $\times$ 40 nm.
The data is extracted from 3 brain areas with neural processes of different morphology and topology.
We process the data to obtain 3 training blocks (512 $\times$ 512 $\times$ 64 voxels), 3 validation blocks (512 $\times$ 512 $\times$ 64 voxels), and 3 test blocks (1250 $\times$ 1250 $\times$ 61 voxels) each with labels and extracted from the respective 3 regions.
Additionally 27 unlabeled blocks of an average size of 1250 $\times$ 1250 $\times$ 64 voxels from each brain area are used for SSL. These blocks were extracted from the original challenge data in a manner that prevents data leakage from the test splits into unsupervised train data. See App.~\ref{sec:appendix_data} for more information about the datasets. 

\subsection{Supervised learning}\label{sec:res_sup}

\begin{figure}[h]
  \centering
  \includegraphics[width=\linewidth]{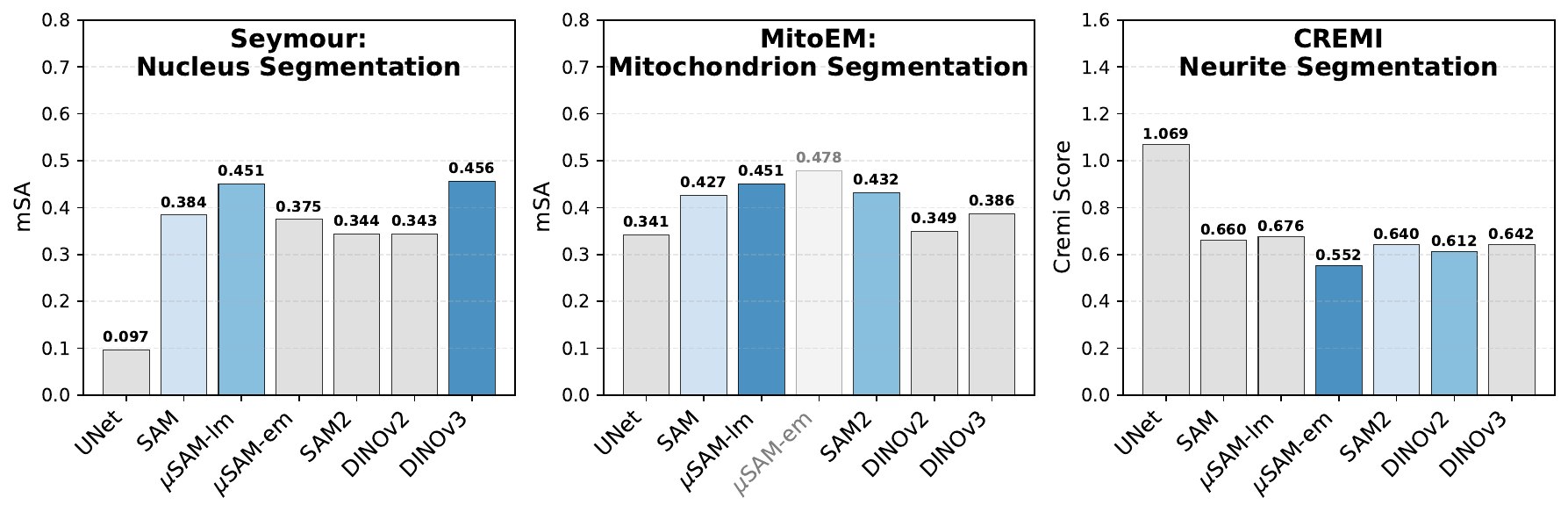}
  \caption{Evaluation of supervised models for nucleus, mitochondrion, and neurite segmentation. We compare a UNet (trained from scratch) with UNETRs initialized with different VFM encoders using mSA (nuclei and mitochondria, higher is better) or CREMI Score (neurites, lower is better). Best three are shaded blue.}
  \label{fig:testset_averages_supervised}
\end{figure}

We compare UNets with UNETRs whose encoders are initialized with weights from different VFMs (Sec.~\ref{sec:supervised}) for the three segmentation tasks (Fig.~\ref{fig:testset_averages_supervised}).

Overall, the UNETR models with SAM-based initialization clearly outperform other models.
For nuclei, the UNet trained from scratch achieves an mSA of 0.097. In contrast, the best-performing UNETR (DINOv3 initialization) reaches an mSA of 0.456.
For mitochondria, the UNet achieves an mSA of 0.341, UNETR ($\mu$SAM-lm) improves its result by 11\%. Note that we don't use $\mu$SAM-em initialization here because this model was trained on the MitoEM dataset. For neurite segmentation, the UNet obtains a CREMI Score of 1.069, while UNETR models again yield substantially better results, with the best-performing one achieving a CREMI Score of 0.552 (lower is better). Overall, models using DINOv2 and DINOv3 backbones tend to show lower performance then SAM-based models. Hence, we exclude them from further experiments.

\subsection{Semi-supervised learning}\label{sec:res_semisup}

\begin{figure}[h!]
\centering
\includegraphics[width=0.85\linewidth]{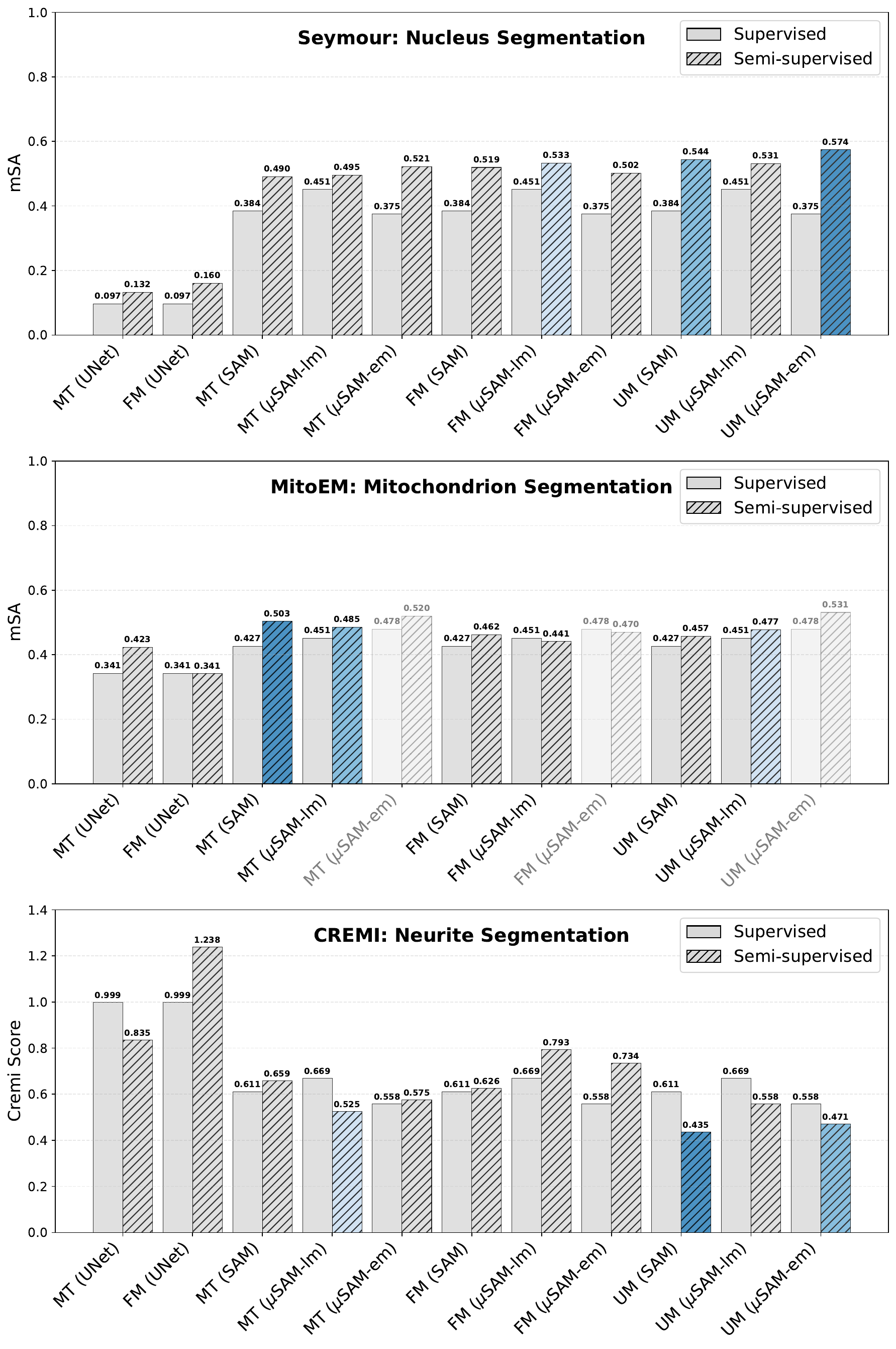}
\caption{Evaluation of SSL for nucleus, mitochondrion, and neurite segmentation, for Mean Teacher (MT), FixMatch (FM), and UniMatch v2 (UM), compared to the respective supervised models (left bars). Base model / encoder initialization is noted in parentheses. For mSA higher is better, for CREMI Score lower is better; three best models are shaded in blue.}
\label{fig:testset_averages}
\end{figure}

For SSL, we evaluate the same segmentation task as before, now also using the unlabeled data for student-teacher-based training, according to the methodology described in Sec.~\ref{sec:semisup}. 
The data used for unsupervised training / validation includes the respective data from the splits used for supervised training / validation, except for the Seymour (nucleus) dataset. 
The results are shown in Fig.~\ref{fig:testset_averages}.

SSL improves the segmentation across settings and datasets. Specifically, semi-supervised configurations with UNETR outperform SSL with a UNet in all cases.
For nucleus segmentation, the UniMatch v2 method with a $\mu$SAM-em backbone achieves the best performance, reaching an mSA of 0.574, representing a substantial improvement compared to the corresponding supervised model with mSA 0.375.
For mitochondria, the Mean Teacher approach with SAM initialization yields the best results, achieving an mSA of 0.503 and an improvement of approximately 7.6\% over the corresponding supervised model and about 16\% compared to the UNet baseline.
For neurite segmentation, UniMatch v2 with SAM initialization performs best, reaching a CREMI Score of 0.435. Further notably here is that, in all cases, FixMatch results in a deterioration in performance compared to the corresponding supervised-trained backbone. 

In addition to the experiments here, we also study the effect of increasing the amount of labeled data (App.~\ref{sec:appendix_data_scaling}), which shows only marginal benefits, the impact of different confidence thresholding strategies (App.~\ref{sec:appendix_conf_thresh}), which are sensitive and depend on the type of target, and perform preliminary experiments for multi-organelle segmentation (App.~\ref{sec:app_multi_organelle}).

\subsection{Domain adaptation}\label{sec:res_da}

\begin{figure}[h]
  \centering
  \includegraphics[width=0.9\linewidth]{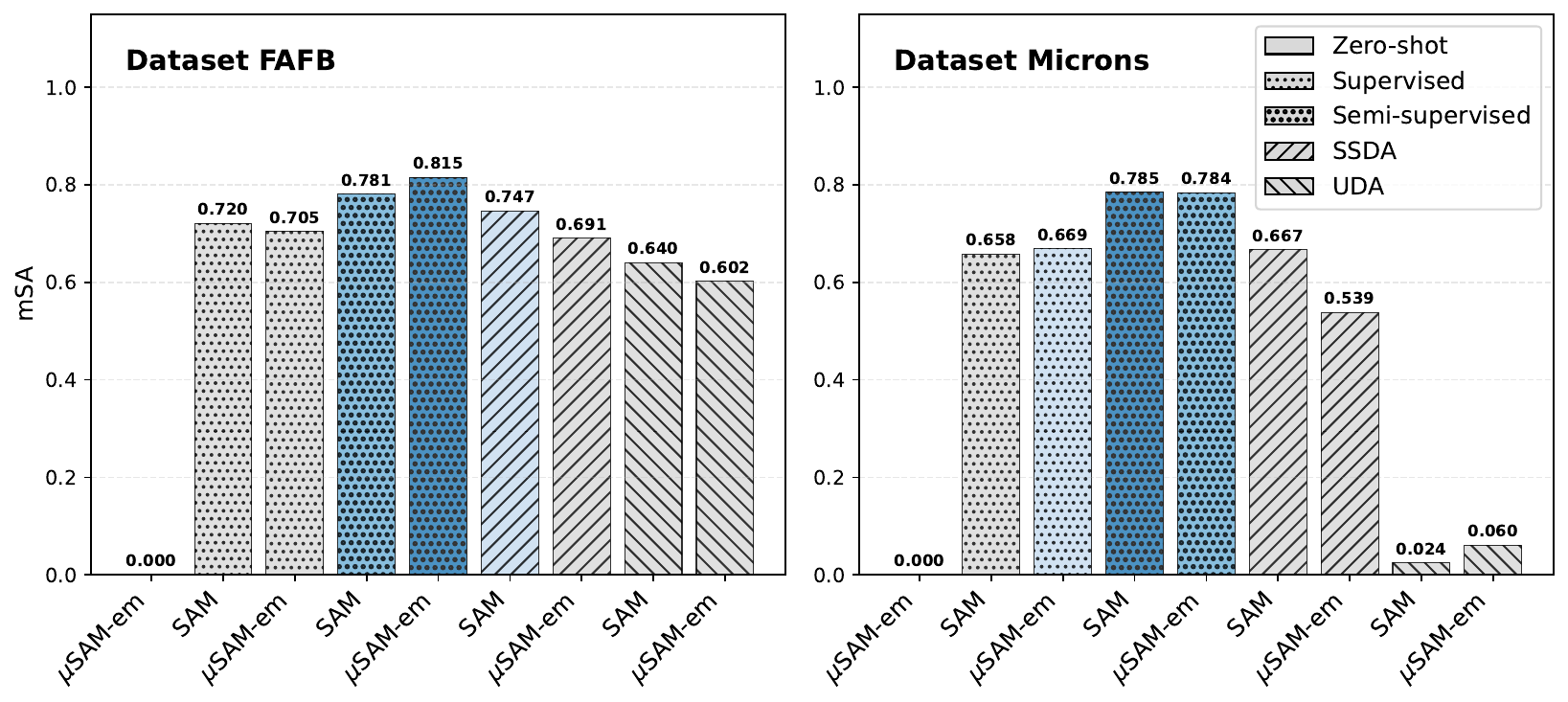}
  \caption{DA for nucleus segmentation (transfer from Seymour to two other datasets) for different training strategies, including supervised learning, SSL, and DA both with (SSDA) and without (SSDA) supervision from labels in the target domain. The leftmost bar corresponds to zero-shot segmentation with $\mu$SAM.}
  \label{fig:nuclei_domain_adaptation}
\end{figure}

\begin{figure}[h]
  \centering
  \includegraphics[width=\linewidth]{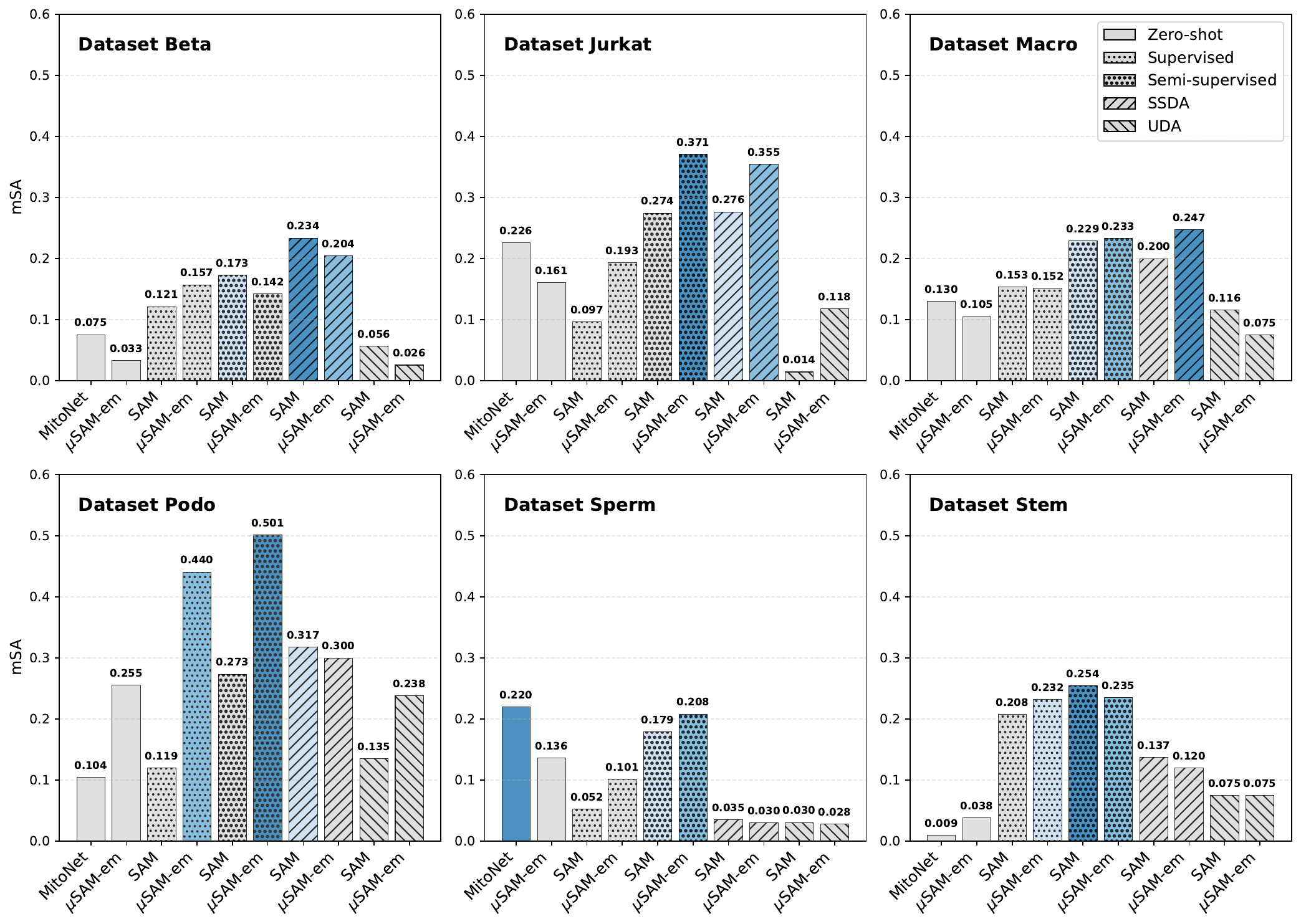}
  \caption{DA for mitochondrion segmentation (transfer from MitoEM to MitoEM v2 datasets) for different training strategies, including supervised learning, SSL, and DA both with (SSDA) and without (UDA) supervision from labels in the target domain. The leftmost bar corresponds to zero-shot segmentation with MitoNet and $\mu$SAM.}
  \label{fig:mito_domain_adaptation}
\end{figure}

For DA experiments, we evaluate nucleus segmentation (Fig.~\ref{fig:nuclei_domain_adaptation}) and mitochondrion segmentation (Fig.~\ref{fig:mito_domain_adaptation}). We compare supervised models, trained only on the labeled train data of the respective target dataset, SSL models, trained on the labeled and unlabeled target data, UDA models, for which the model is pre-trained on the fully labeled source dataset and that are then trained on the target using only the unsupervised loss, and SSDA, which additionally uses the labeled blocks of the target domain for SSL. See also Sec.~\ref{sec:da} for the DA methodology. 

For DA in nucleus segmentation, we study the transfer from Seymour to two other datasets, FAFB and MICrONS (Sec.~\ref{sec:data}). Here, SSL trained directly on the target domain, without prior training on the Seymour data, achieve the best performance. Domain adaptation to FAFB is substantially more successful than adaptation to MICrONS, which suggests that the FAFB dataset is more similar to Seymour in terms of imaging characteristics and data distribution than MICrONS. However, in both cases DA does not offer an advantage compared to SSL, indicating a domain shift too large to profit with our method. Here, we also include results from the pretrained $\mu$SAM-em model, which however does not yield meaningful results.
We evaluate DA for nucleus segmentation for further, more diverse datasets (from whole-body EM of the \textit{Drosophila} \cite{peale2024igor} and \textit{Platynereis} larva \cite{platy}) in App.~\ref{sec:app_nucleus_da}.

A similar trend is observed for mitochondrion segmentation. Here, pure SSL on the target domain again shows strong performance. For Beta and Macro datasets, SSDA yields the best results, indicating a smaller distribution shift to MitoEM v1 compared to the other target domains. We also evaluate MitoNet and $\mu$SAM in a zero-shot setting. For the Sperm dataset, MitoNet achieves the best performance. For the other datasets SSL and DA outperform both MitoNet and $\mu$SAM.

\subsection{Conclusion}
We systematically evaluate supervised learning, SSL, and DA for instance segmentation of nuclei, mitochondria, and neurites in large-scale EM data. Our results show that supervised learning profits massively from initialization with VFMs. Specifically, we find that using SAM-based models provides a substantial benefit, with minor and inconsistent differences between the original SAM and more domain-specific versions ($\mu$SAM-lm/em). In contrast, DINOv2/v3 do not provide a benefit. 

Furthermore, we find that student-teacher-based SSL provides substantial benefits for all segmentation tasks.
Overall, FixMatch performs worse than the other two methods. Mean Teacher offers a balanced trade-off between robustness, computational cost and performance, and is therefore the recommended standard method. However, if sufficient computational resources are available, UniMatch v2 is preferable, as it achieves the best results in several scenarios, although its more complex scheme comprising three extended views and complementary feature dropout does not offer consistent advantages across all tasks.

The DA experiments reinforce the value of SSL, which performed best in most settings. They further showed the need for our methodology, as the pre-trained models for mitochondrion and nucleus segmentation, MitoNet and $\mu$SAM, performed poorly in a zero-shot setting. However, they showed that DA suffers from large domain shifts in our setting: SSDA improved over SSL in only two cases and UDA generally performed poorly. Note that other work has found advantages of student-teacher-based (U)DA for smaller domain shifts in EM, e.g. \cite{muth_synapsenet_2025, archit_probabilistic_2023}. Thus, we believe that our DA methodology will prove beneficial in such cases. 

Future work could address remaining limitations by further investigating confidence thresholding strategies and pseudo-label generation. Further possible directions are alternative model outputs for instance segmentation, such as distance-based representations, and pseudo-labels derived only from foreground prediction, which is generally more robust than other targets. Another possibility is to threshold teacher predictions to obtain binary labels rather than continuous ones as in our current set-up, to possibly avoid sensitivity in confidence thresholding. Additional, the inconsistent trend when adding further labeled data in SSL should be investigated.
Finally, the development of better and more general EM-specific foundation models holds promise and could further elevate our contribution by better model initialization. Our methodology could also be directly integrated within tools built around microscopy foundation models, such as $\mu$SAM, to improve fine-tuning on user-annotated data.

\FloatBarrier

%
%
\bibliographystyle{splncs04}
\bibliography{main}

\newpage
\section*{Appendix}

\subsection{Metrics}\label{sec:appendix_metrics}

mSA is calculated as the mean of segmentation accuracies over multiple overlap thresholds in the range from 0.5 to 0.95 with step size 0.05. The SA represents the ratio of correctly predicted instances to the total number of predicted and ground truth instances:
\begin{align*}
    SA = \frac{TP}{TP + FP + FN},
\end{align*} 
where TP, FP, and FN denote the number of true positive, false positive, and false negatives, according to the selected overlap threshold.

The CREMI Score is defined as the geometric mean of two complementary measures, the sum of the Variation of Information (VI) \cite{meilua2003comparing}, and the Adapted Rand Error (ARAND) \cite{hubert1985comparing}:
\begin{align*}
    CREMI\_Score = \sqrt{ARAND \cdot (VI_S + VI_M)}.
\end{align*} 
The VI is decomposed into a split term $VI_S$ and a merge term $VI_M$. The split term penalizes over-segmentation, where a single ground truth object is divided into multiple predicted segments, while the merge term penalizes under-segmentation, where multiple ground truth objects are incorrectly merged into a single predicted segment. The ARAND component measures the agreement between pairs of voxels in the predicted and ground-truth segmentation. It captures both split and merge errors from a pairwise perspective. Lower values of the CREMI Score indicate better segmentation performance, reflecting fewer over-segmentation and under-segmentation errors.

\subsection{Data}\label{sec:appendix_data}

An overview of all datasets used for the main experiments is given in Sec.~\ref{sec:data}. Briefly, we use three datasets for nucleus segmentation: Seymour, FAFB, and MICRONS. The latter two are used for the DA experiments. For mitochondrion segmentation we use the MitoEM dataset and, for DA, six out of eight datasets from MitoEM v2. An overview of the datasets from MitoEM v2 is given in Tab.~\ref{tab:me2_datasets}.
We use the CREMI dataset for neurite segmentation.

We use additional datasets for the experiments in the appendix. For the extended experiments on DA for nucleus segmentation we use the Platynereis and the Igor dataset. The first dataset is derived from a \textit{Platynereis} larva imaged in block face EM. We use the data at a voxel size of 80 $\times$ 80 $\times$ 100 nm and use one block for supervised training, one for supervised validation, two for testing and eight additional blocks for SSL. The blocks measure 375 $\times$ 375 $\times$ 100 voxels, the patch shape is accordingly set to 375 in x and y dimension. The second dataset, Igor, is derived from a \textit{Drosophila} larva \cite{peale2024igor} imaged in ssTEM. We use the data at a voxel size of 40 $\times$ 40 $\times$ 35 nm, with one block for supervised training and validation, respectively; 4 test blocks and 56 additional blocks without labels for SSL (each block of size 512 $\times$ 512 $\times$ 64 voxels). The annotations for this dataset were created by us using interactive segmentation in $\mu$SAM.
These two datasets cover the whole body, containing more diverse tissues compared to the two datasets used in Fig.~\ref{fig:nuclei_domain_adaptation}, which are both derived from neural tissue.
These experiments are summarized in App.~\ref{sec:app_nucleus_da}.

In addition, we perform preliminary experiments on general purpose organelle segmentation using data from OpenOrganelle \cite{openorganelle} (App.~\ref{sec:app_multi_organelle}).
Here, we use investigate sematic segmentation of endoplasmic reticulum (ER), vesicles, and endosomes, using labels provided by OpenOrganelle.
The data itself comes from different cells and tissues, imaged with FIBSEM. We use data at an isotropic resolution of 8 nm, using 62 / 9 blocks for supervised training / validation with an additional 14 blocks for SSL for ER, 38 / 5 / 14 for vesicles, and 57 / 7 / 14 for endosomes. The blocks have different shapes, blocks without labels used for SSL are substantially larger than the blocks used for supervised learning, resulting in a larger dataset for SSL despite the lower number of blocks. 

\begin{table}[ht]
\centering
\scriptsize
\begin{tabular}{l l l l l l r c r c}
\toprule
Name & Organism & Tissue & Cell Type & Modality & Avg Shape & Voxels & Split & \#Mito \\
\midrule
Beta   & Mouse       & Pancreas   & $\beta$-cell     & FIB-SEM & (874,669,979)   & 3961M & 3/1/3  & 1935 \\
Jurkat & Human       & Cell line  & Jurkat          & FIB-SEM & (1024,1024,256) & 805M  & 1/1/1  & 372  \\
Macro  & Human       & Cell line  & Macrophages     & FIB-SEM & (1024,1024,256) & 805M  & 1/1/1  & 497  \\
Podo   & Mouse       & Kidney     & Podocyte        & FIB-SEM & (1024,1024,256) & 805M  & 1/1/1  & 1333 \\
Sperm  & Drosophila  & Gonad      & Spermatocyte    & FIB-SEM & (1024,1024,341) & 1073M & 3/1/4  & 775  \\
Stem   & Mouse       & Brainstem  & Mix             & SBF-SEM & (1024,1024,100) & 300M  & 1/1/1  & 2058 \\
\bottomrule
\end{tabular}
\caption{Overview of MitoEM v2 datasets and their properties \cite{liu2025mitoemv2}.}
\label{tab:me2_datasets}
\end{table}

\subsection{Training Parameters}

\begin{table}[ht]
\centering
\begin{tabular}{l l l l r r l}
\hline
Model & Backbone & Model type & patch shape & batch size \\
\hline
UNet & - & - & 512, 512, 24 & 2  \\
SAM & SAM & vit\_b & 512, 512, 4 & 1 \\
$\mu$SAM-lm & SAM & vit\_b\_lm & 512, 512, 4 & 1 \\
$\mu$SAM-em & SAM & vit\_b\_em & 512, 512, 4 & 1 \\
SAM2 & SAM2 & hvit\_b & 512, 512, 4 & 2 \\
DINOv2 & DINOv2 & vit\_b & 512, 512, 4 & 2 \\
DINOv3 & DINOv3 & vit\_b & 512, 512, 4 & 2 \\
\hline
\end{tabular}
\caption{Overview of model backbones and training parameters}
\label{tab:training_parameters}
\end{table}

\FloatBarrier
\subsection{Training data scaling} \label{sec:appendix_data_scaling}

As referenced in \ref{sec:semisup}, we also trained the best model set-up on each dataset with an increasing the amount of labeled data. The number of labeled training blocks is increased by factors of 2, 3, and 4 relative to the base configuration. All other training hyperparameters were not changed to ensure comparability. 

\begin{figure}[h!]
  \centering
  \includegraphics[width=\linewidth]{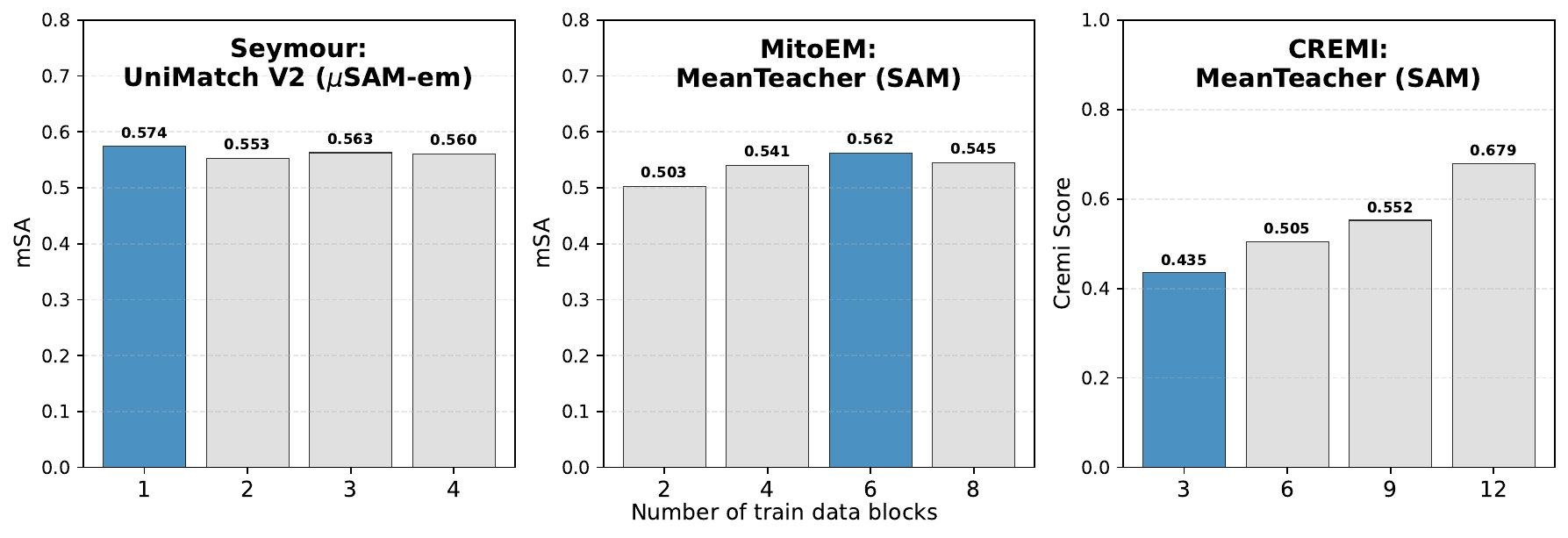}
  \caption{Evaluation of SSL frameworks for nucleus, mitochondrion, and neurite segmentation with an increasing number of labeled training data blocks. Best result is marked in blue.}
  \label{fig:data_scaling}
\end{figure}

The training with an increasing number of supervised data blocks produces inconsistent results across different tasks. The individual results are displayed in Fig.~ \ref{fig:data_scaling}. 
For nucleus segmentation on the Seymour dataset, using a single supervised training block achieves the best performance, but using additional blocks results in only slightly lower performance. For mitochondrion segmentation, the best results are obtained using six supervised training blocks, suggesting that this task benefits from a moderate increase in labeled data. In contrast, for neurite segmentation, the model with the fewest labeled training blocks achieves the best performance. 
One possible explanation for these observations is that adding more supervised training blocks increases data variability, leading to less confident model predictions. This effect can be problematic in semi-supervised settings that rely on confidence thresholding, as increased uncertainty in teacher predictions can reduce the quality or quantity of pseudo-labels used for unsupervised training.

\FloatBarrier
\subsection{Confidence threshold scheduling} \label{sec:appendix_conf_thresh}

To further investigate the confidence threshold strategy, which we identify as the most sensitive component of the semi-supervised pipeline, we evaluate six different thresholding configurations using the Mean Teacher, FixMatch, and UniMatch v2 frameworks for SSL based on the SAM encoder as before. 

In setting (1) - (3), the confidence threshold $t_c$ is fixed to a constant value of 0.5, 0.7, and 0.9, respectively. In (4), after a warm-up phase of setting $t_c = 0.5$ for 1,000 iterations, the threshold is set to 0.9 and subsequently reduced if the validation loss stagnates. Finally, in setting (5) $t_c$ is increased stepwise from 0.5 to 0.9 (step size 0.05) over the first half of the training, and in (6) over the entire training duration.
All experiments are done on the MitoEM dataset, with the model predicting foreground and boundary output channels, and the CREMI dataset where the model outputs affinities for x, y and z dimensions as described in Sec.~ \ref{sec:supervised}.

\begin{figure}[h!]
  \centering
  \includegraphics[width=\linewidth]{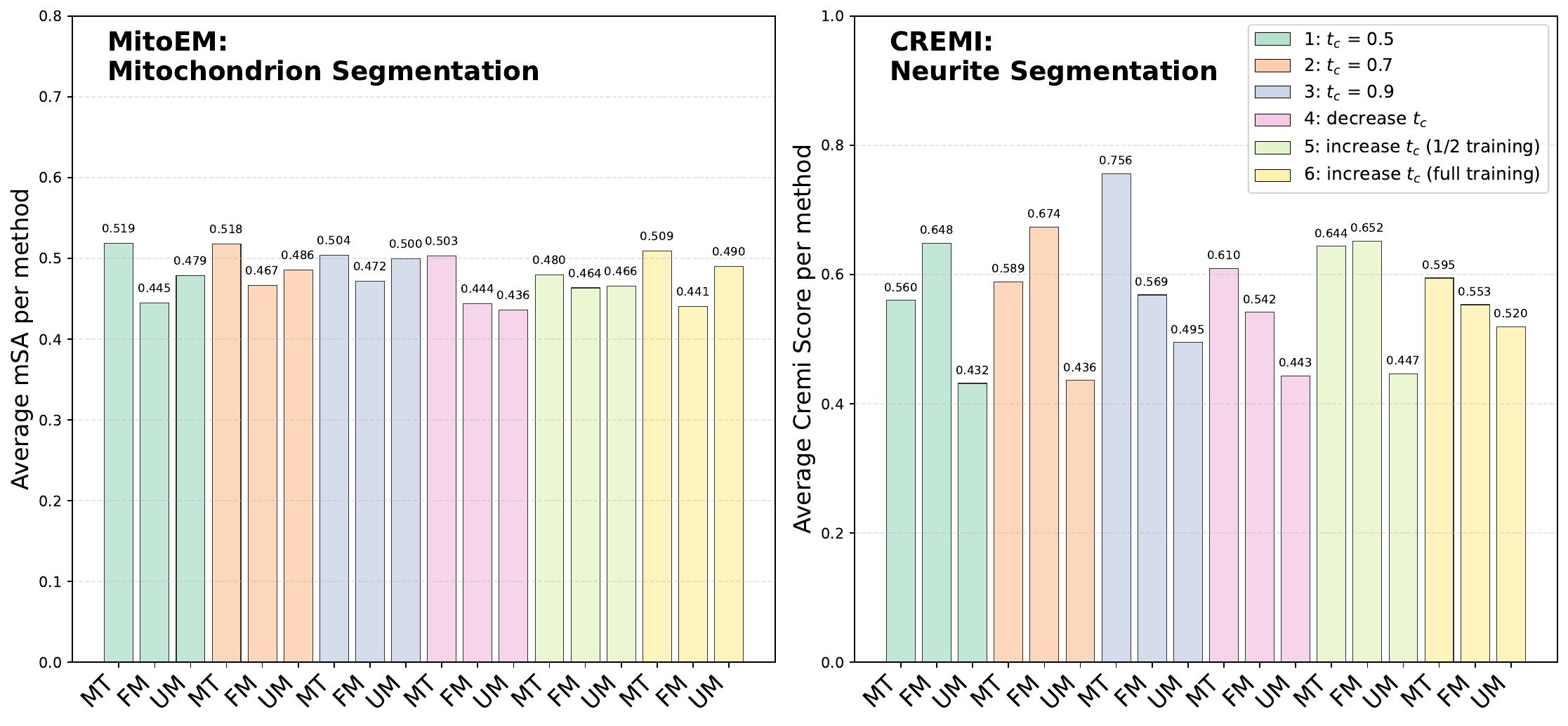}
  \caption{Evaluation per confidence threshold strategy and SSL framework Mean Teacher, FixMatch and UniMatch v2; bars are colored according to the confidence threshold setting.}
  \label{fig:ct_models}
\end{figure}

For the MitoEM dataset, we do not observe significant differences between the evaluated confidence thresholding strategies, with a fixed threshold of $t_c = 0.9$ yielding the best overall results. In contrast, the CREMI dataset shows a stronger dependence on the threshold schedule. Here, the confidence threshold needs to be set lower at least during the early stages of training to avoid filtering out uncertain predictions entirely, which would otherwise prevent effective learning from unlabeled data. Moreover, gradually increasing the threshold proves beneficial, with slower schedules outperforming strategies that raise the threshold to 0.9 within the first half of the training. Additionally, Figure \ref{fig:ct_models} highlights again the advantage of using UniMatch v2 in this setting as it shows superior performance across all confidence threshold strategies compared to Mean Teacher and FixMatch. 

\begin{figure}[h!]
  \centering
  \includegraphics[width=\linewidth]{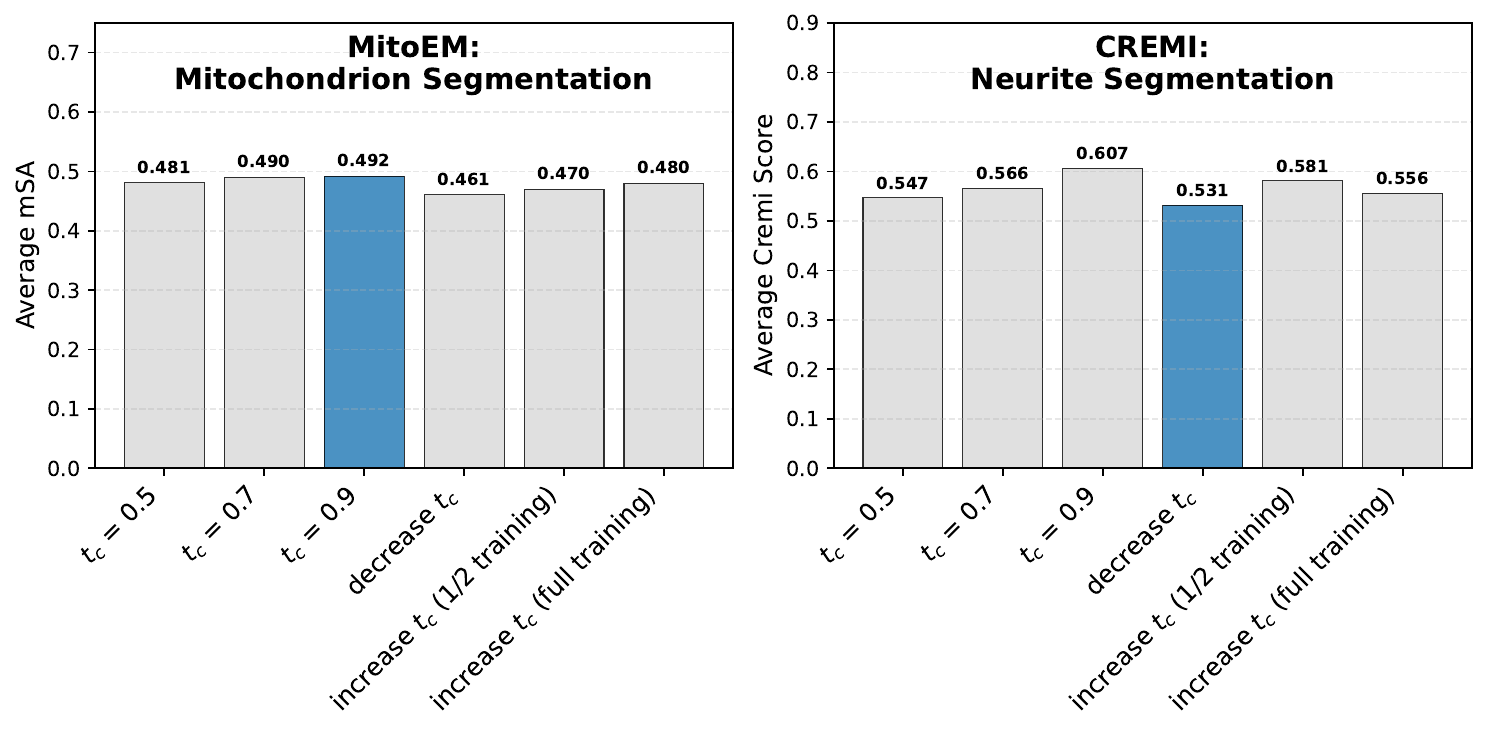}
  \caption{Evaluation per confidence threshold strategy for the three different SSL frameworks Mean Teacher, FixMatch and UniMatch v2; best result is marked in blue.}
  \label{fig:ct_average}
\end{figure}

\FloatBarrier
\subsection{Extended nucleus domain adaptation} \label{sec:app_nucleus_da}

For DA experiments, we further evaluate nucleus segmentation (Fig.~\ref{fig:nuclei_domain_adaptation}) also on two full body EM volumes (1st instar Drosophila larva \cite{peale2024igor} and \textit{Platynereis} larva \cite{platy}, see Sec.~\ref{sec:appendix_data}). As before, we compare supervised models, trained only on the labeled train data of the respective target dataset, SSL models, trained on the labeled and unlabeled target data, UDA models, for which the model is pre-trained on the fully labeled source dataset and that are then trained on the target using only the unsupervised loss, and SSDA, which additionally uses the labeled blocks of the target domain for SSL. See also Sec.~\ref{sec:da} for the DA methodology. To make sure, that maybe less confident predictions on the full body volumes are not filtered out by a too high confidence threshold, we set the threshold to 0.5 for all these experiments and also re-evaluate the results with that setting on the FAFB and MICrONS dataset. 

For FAFB and MICrONS, as already discussed in Section~\ref{sec:res_da}, SSL settings achieve the best performance. Domain adaptation to the FAFB dataset also yields strong results, which is likely due to the close similarity between FAFB and the Seymour brain data in terms of structure and imaging characteristics.
In contrast, the adaptation to full-body datasets becomes substantially more challenging. In these settings, purely supervised training on the target data yields the best performance. SSL performs significantly worse for the Igor dataset, which is likely caused by poor-quality pseudo-labels that prevent effective learning. Similarly, DA to the Platy dataset does not lead to meaningful improvements, suggesting that the source and target domains are too dissimilar for effective transfer.
These results re-iterate that the success of domain adaptation with our method strongly depends on the domain shift.

\begin{figure}[h!]
  \centering
  \includegraphics[width=\linewidth]{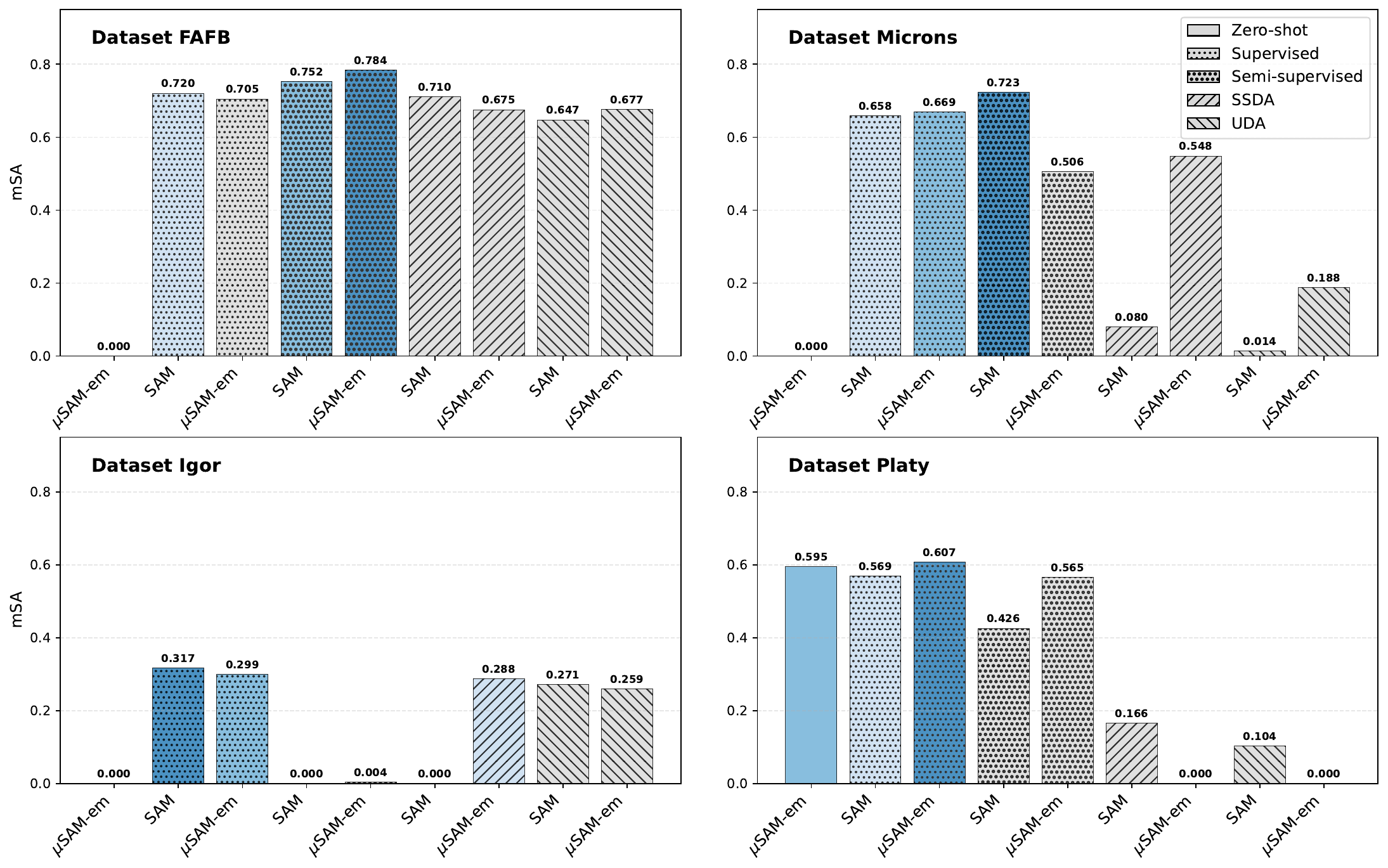}
  \caption{DA for nucleus segmentation (transfer from Seymour to four other datasets, including the full-body volume Igor and Platynereis) for different training strategies, including supervised learning, SSL, and DA both with (SSDA) and without (SSDA) supervision from labels in the target domain. In all SSL and DA settings, $t_c = 0.5$ is used. The leftmost bar corresponds to zero-shot segmentation with $\mu$SAM.}
\label{fig:da_all_nuclei}
\end{figure}

\FloatBarrier
\subsection{Organelle Segmentation} \label{sec:app_multi_organelle}

\begin{figure}[h!]
  \centering
  \includegraphics[width=\linewidth]{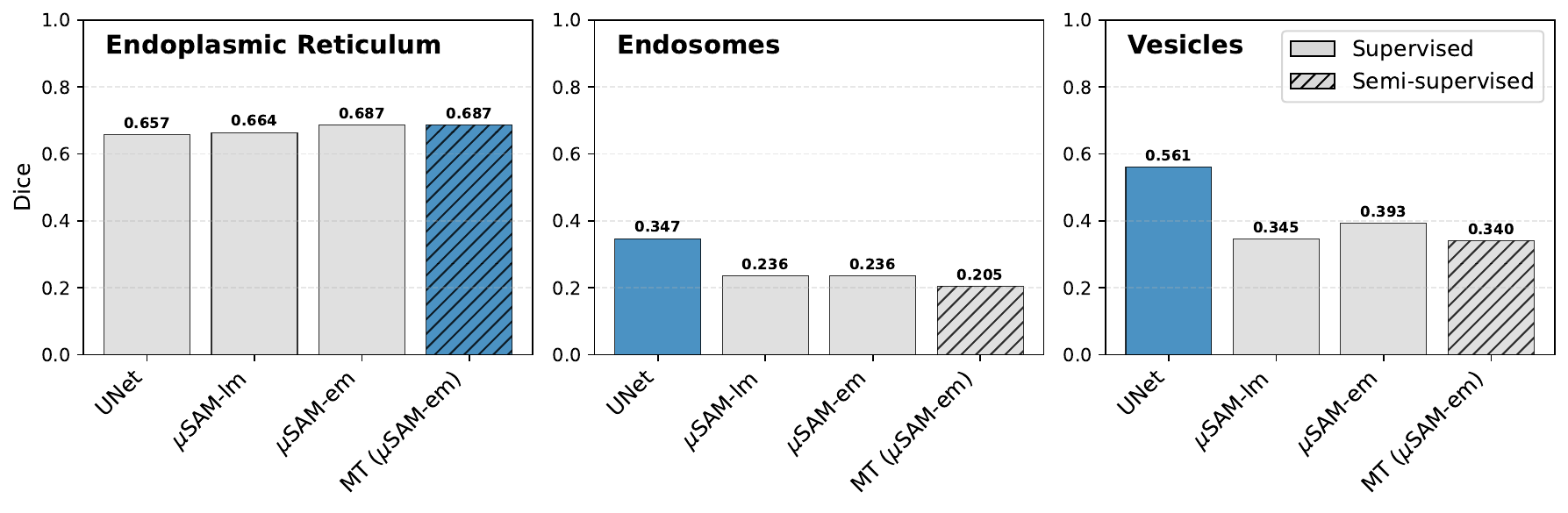}
  \caption{Organelle segmentation results. Individual models were trained for the three binary segmentation tasks: ER, endosome and vesicle segmentation. We trained a UNet and UNETR models initialized with $\mu$SAM weights (supervised) as well as a Mean Teacher using the best performing UNETR $\mu$SAM-em as base model. The results are evaluated with the Dice score (higher is better) and the best result is marked in blue.}
\label{fig:organelle-seg}
\end{figure}

We perform preliminary experiments on organelle segmentation, specifically for ER, vesicle, and endosome segmentation, using data from OpenOrganelle, see App.~\ref{sec:appendix_data} for details. Here, we compare UNETR models initialized with $\mu$SAM-lm / em model with a UNet trained from scratch (supervised setting) and with a mean teacher based on the best UNETR model for the respective task. We train individual models for each task and train them for foreground / background segmentation with a Dice loss, using default training parameters otherwise. The results are shown in Fig.~\ref{fig:organelle-seg}.

Here, we see a different trend compared to all other results where we compared supervised learning with a UNet (randomly initialized weights) and a UNETR initialized with foundation model weights, where the UNETR models outperformed the UNet (cf. Fig.~\ref{fig:testset_averages_supervised} for those results): the UNETR performs on par (ER) or substantially worse (vesicles, endosomes) than the UNet. We believe that this observation can be attributed to two factors: the UNETR models were trained with a smaller patch shape in z (depth axis) compared to the UNet due to memory constraints: they were trained with a patch shape of 128 $\times$ 128 $\times$ 8, compared to 64 $\times$ 64 $\times$ 64 for the UNet. While we also had similar discrepancies in the patch shape across z for other training tasks (see Tab.~\ref{tab:training_parameters}), the corresponding data had a lower voxel size, so that these models still had a sufficient physical field of view. Conversely, the structures segmented here are smaller and have a more filigree structure, thus requiring the larger voxel size. Furthermore, their smaller size and structure also makes exact spatial resolution more important, which may disadvantage the patching based vision transformers.
We also don't see a big advantage due to SSL with a mean teacher, likely caused by similar effects that contribute to the bad performance of the supervised UNETR models.

Overall, these results show the need for further research to translate the clear benefits we have seen for other segmentation tasks to small or finely structured organelles.

\FloatBarrier
\subsection{Qualitative Results} \label{sec:app_quali_results}

\begin{figure}[h!]
  \centering
  \includegraphics[width=\linewidth]{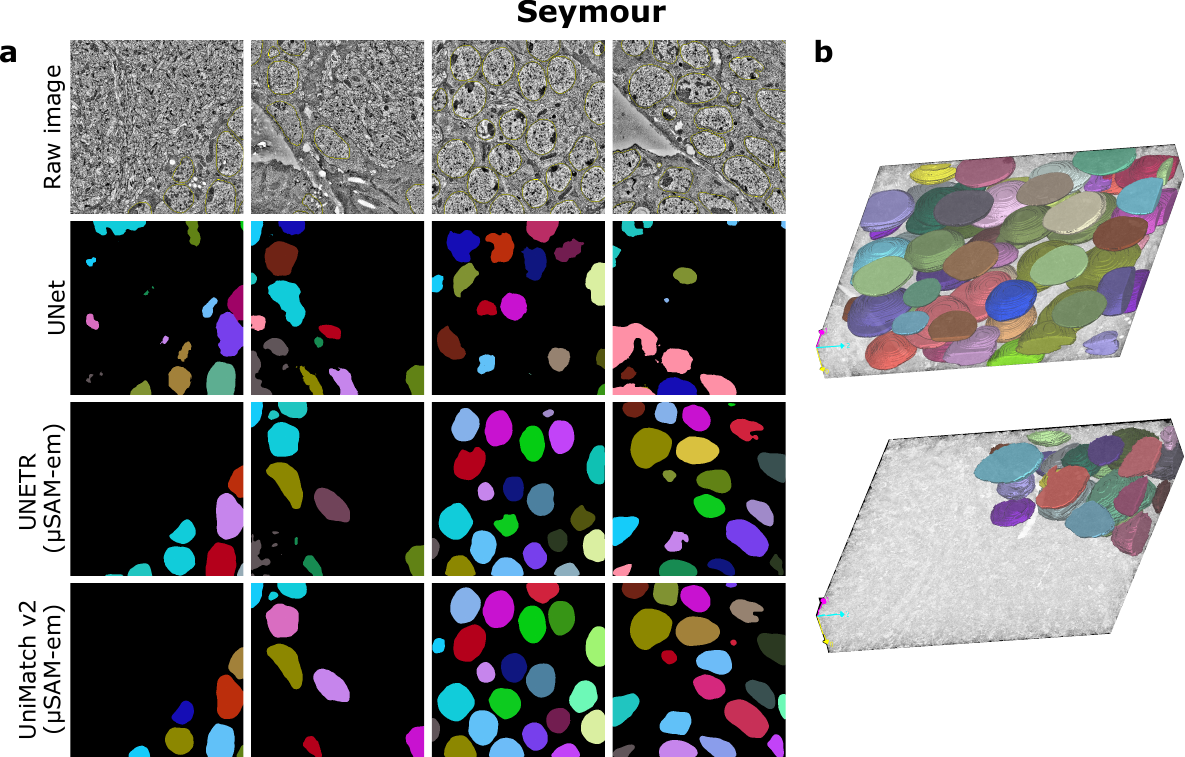}
  \caption{\textbf{a)} Prediction comparisons of UNet, UNETR ($\mu$SAM-em), UniMatch v2 ($\mu$SAM-em) on the different Seymour test data blocks, \textbf{b)} two selected 3D views of the test data with the UniMatch v2 prediction.}
  \label{fig:pred_seymour}
\end{figure}

\begin{figure}[h!]
  \centering
  \includegraphics[width=\linewidth]{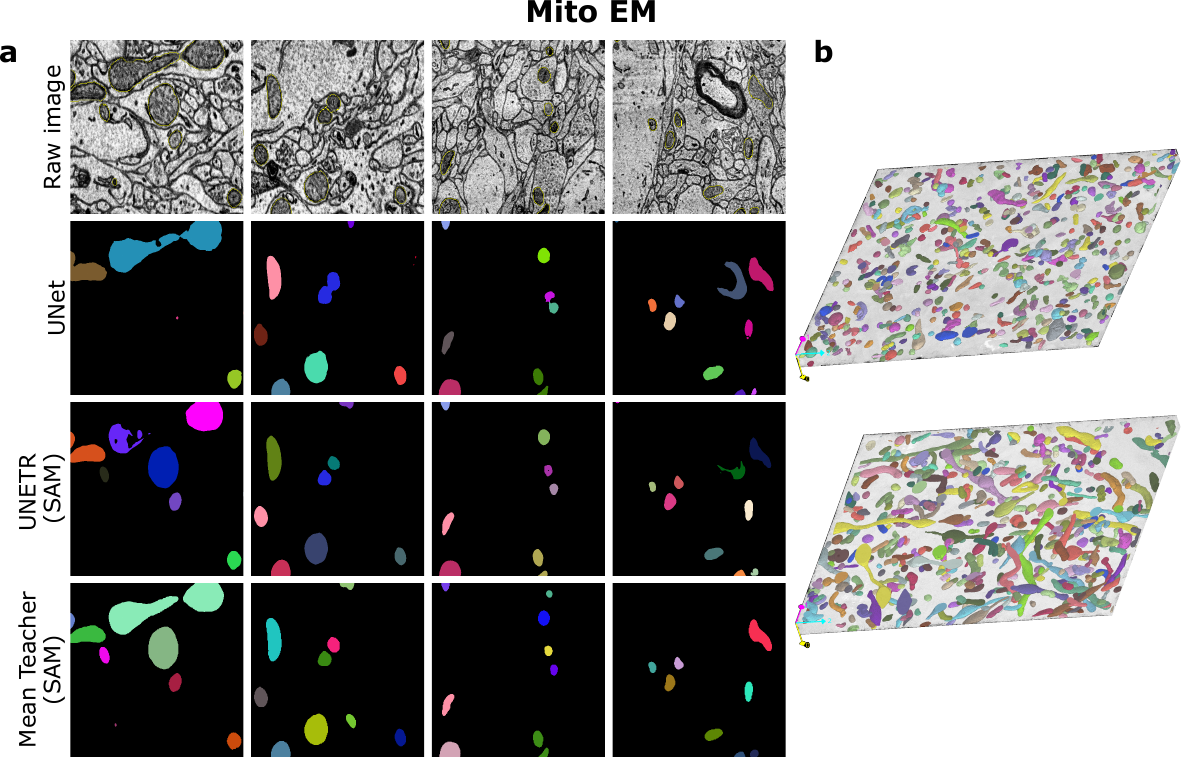}
  \caption{\textbf{a)} Prediction comparisons UNet, UNETR (SAM), Mean Teacher (SAM) on the different Mito EM test data blocks, \textbf{b)} two selected 3D views of the human and rat test data with the Mean Teacher prediction.}
  \label{fig:pred_mito}
\end{figure}

\begin{figure}[h!]
  \centering
  \includegraphics[width=\linewidth]{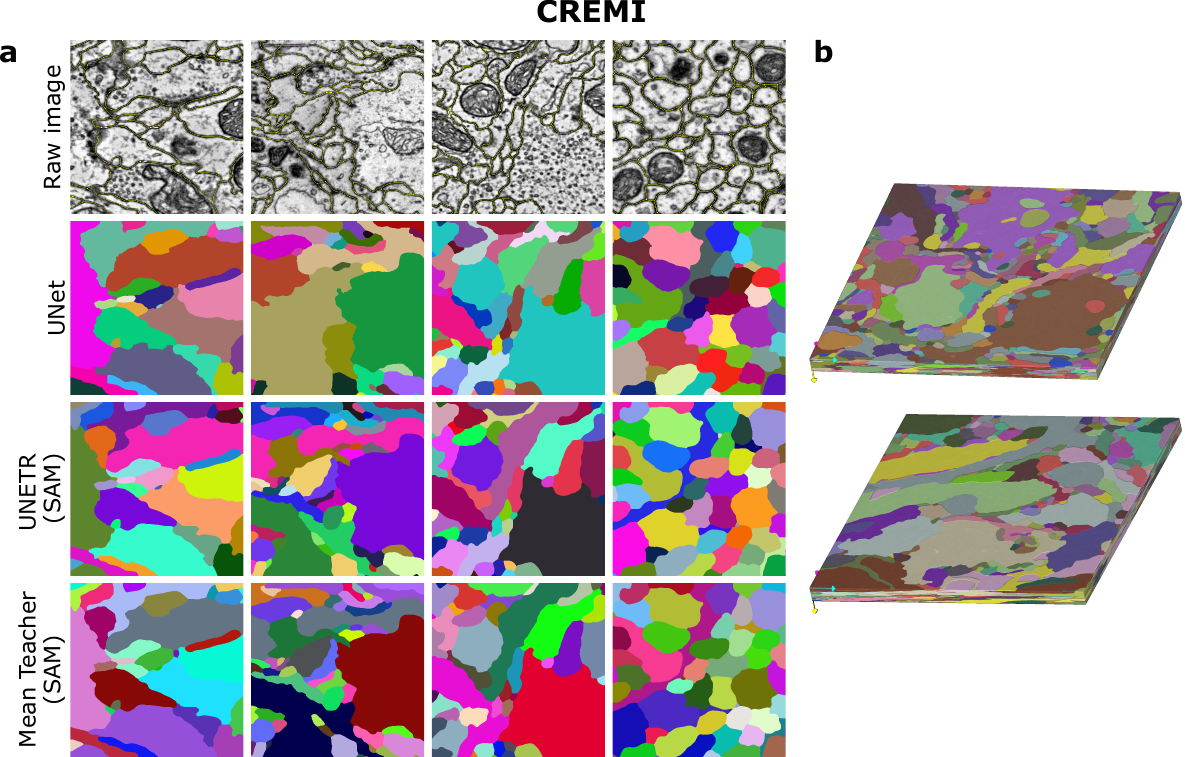}
  \caption{\textbf{a)} Prediction comparisons UNet, UNETR (SAM), Mean Teacher (SAM) on the different CREMI test data blocks, \textbf{b)} two selected 3D views of the test data with the Mean Teacher prediction.}
  \label{fig:pred_cremi}
\end{figure}

Figs.~\ref{fig:pred_seymour}–\ref{fig:pred_cremi}a) show the inferred instance segmentation produced by the best-performing semi-supervised model configurations described in Section~\ref{sec:res_semisup}, together with the corresponding underlying supervised trained UNETR model and the UNet trained from scratch. 

Overall, the UNet predictions show a greater number of merged and fragmented objects across all datasets. In contrast, UNETR and the SSL frameworks are better at detecting details and finer structures, and are less affected by image artefacts.
This behavior is particularly visible for the Seymour nuclei dataset (Fig.~\ref{fig:pred_seymour}a), second row), where the UNet predicts nuclei in regions where none are present (1st image), and shows multiple merged nuclei (last image). In contrast, the UNETR models based on the $\mu$SAM-em encoder perform substantially better, although a small number of errors, as fragmented objects (1st image) remain that are no longer present in the UniMatch v2 segmentation. A similar pattern is observed for the MitoEM dataset, where, for example, only the Mean Teacher model correctly avoids predicting the dark circular structure (Fig.~\ref{fig:pred_mito}a), fourth column) as a mitochondrion.

In neurite segmentation on the CREMI dataset, the UNet's main weakness is its tendency to produce large merge errors. These errors are particularly harmful for connectivity analysis because merged neurons destroy the underlying topology more severely than over-segmentation does. This issue is particularly evident in the UNet predictions (\ref{fig:pred_cremi}a), second row, first and second images), where large merge errors occur along the left border of the image. By contrast, UNETR- and Mean-Teacher-based models recover finer details and more delicate neurite structures.

For all datasets, Fig.~\ref{fig:pred_seymour}–\ref{fig:pred_cremi}b) additionally show two test blocks or sub-volumes rendered in 3D, providing a more comprehensive overview of the spatial distribution and localization of segmented objects within the volumes.

\end{document}